\documentclass[letterpaper, 10 pt, conference]{IEEEtran}  

\IEEEoverridecommandlockouts                              

\usepackage{cite}
\usepackage{amsmath,amssymb,amsfonts}
\usepackage{graphicx}
\usepackage{subcaption} 
\usepackage{textcomp}
\usepackage{xcolor}
\usepackage{algorithm}
\usepackage{algpseudocode}
\usepackage{multirow}
\usepackage{booktabs}
\usepackage{hyperref}

\usepackage{gensymb}
\usepackage{float}

\graphicspath{{figures}}
\algrenewcommand\algorithmicwhile{\textbf{while}}
\algrenewcommand\algorithmicif{\textbf{if}}
\algrenewcommand\algorithmicfor{\textbf{for}}
\algrenewcommand\algorithmicreturn{\textbf{return}}

\newenvironment{smallcases}{\left\{\begin{smallmatrix}}{\end{smallmatrix}\right.}
\title{GSeg3D: A High-Precision Grid-Based Algorithm for Safety-Critical Ground Segmentation in LiDAR Point Clouds}

\author{
\IEEEauthorblockN{Muhammad Haider Khan Lodhi*\thanks{This work has been carried out within the project Robdekon2 (grant number: 13N16537), 
funded by the German Federal Ministry of Education and Research (BMBF). All code will be published after acceptance.}}

\IEEEauthorblockA{
German Research Center for Artificial Intelligence (DFKI)\\
Robotics Innovation Center\\
Bremen, Germany \\
Email: Haider\_Khan.Lodhi@dfki.de \\
ORCID: \href{https://orcid.org/0009-0008-1199-3489}{0009-0008-1199-3489}
}
*Corresponding author
~\\
\and
\IEEEauthorblockN{Christoph Hertzberg}
\IEEEauthorblockA{
German Research Center for Artificial Intelligence (DFKI)\\
Robotics Innovation Center\\
Bremen, Germany\\
Email: Christoph.Hertzberg@dfki.de \\
ORCID: \href{https://orcid.org/0009-0005-6467-0286}{0009-0005-6467-0286}
}
}

\begin{document}
\maketitle

\begingroup
\renewcommand\thefootnote{}
\footnotetext{
© 2025 IEEE. Personal use of this material is permitted.

This is the author's accepted manuscript of the paper published in:
\textit{2025 7th International Conference on Robotics and Computer Vision (ICRCV)}.

The final published version is available at:
https://ieeexplore.ieee.org/document/11349133
}
\addtocounter{footnote}{-1}
\endgroup

\begin{abstract}
Ground segmentation in point cloud data is the process of separating ground points from non-ground points. 
This task is fundamental for perception in autonomous driving and robotics, where safety and reliable operation 
depend on the precise detection of obstacles and navigable surfaces. Existing methods often fall short of the 
high precision required in safety-critical environments, leading to false detections that can compromise 
decision-making. In this work, we present a ground segmentation approach designed to deliver consistently 
high precision, supporting the stringent requirements of autonomous vehicles and robotic systems operating 
in real-world, safety-critical scenarios.
\end{abstract}

\begin{IEEEkeywords}
Ground Segmentation, Grid-Based Segmentation, LiDAR, Robotics, Autonomous Driving.
\end{IEEEkeywords}

\section{Introduction}
The perception system of an autonomous vehicle relies on multiple sensors to capture critical features of the 
surrounding environment. Among these, LiDAR has become a core technology, providing dense, high-resolution 
point clouds that enable detailed spatial understanding \cite{Gomes2023}. A single LiDAR scan can contain 
over a million points, capturing both ground and non-ground surfaces within the region of interest. However, 
not all points are equally relevant for downstream perception tasks.

For example, object and obstacle detection algorithms often misclassify ground points as obstacles, leading to 
false positives that can compromise safety and increase computational load \cite{Firkat2023}. Accurate ground segmentation 
is essential for reliable perception, traversability analysis, 
navigation, and map generation in autonomous systems \cite{TraversabilitySurvey, tsaiground, Arora2023, Jimenez2021, Quatro}.

\begin{figure}[b]
\centering
\includegraphics[width=0.7\linewidth]{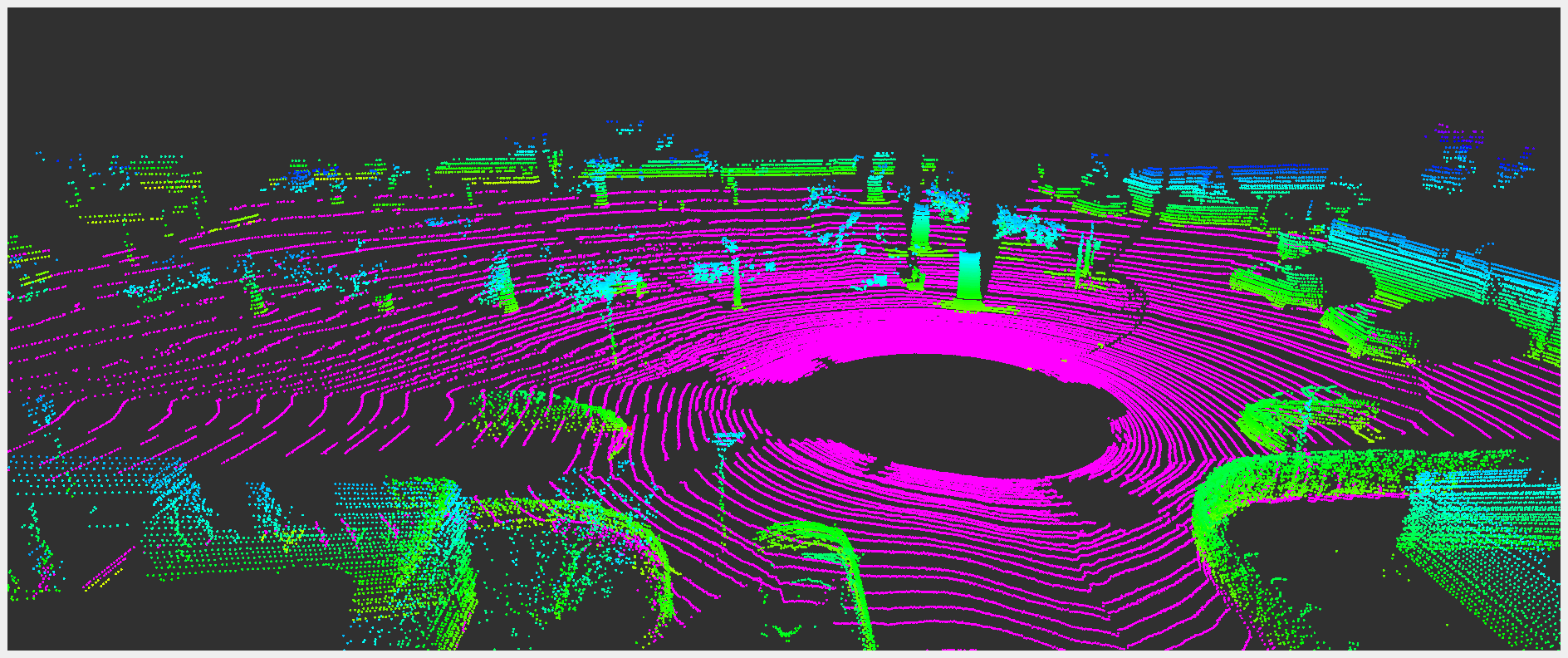}
\caption{GSeg3D: The point cloud is segmented into ground points (magenta) and non-ground points, which are color-coded according to 
their height to visualize vertical structures.
}
\label{fig:results}
\end{figure}

In this work, we address the need for high-precision ground segmentation in LiDAR point clouds which is a critical requirement for 
safety-critical applications in autonomous driving and robotics. We present GSeg3D, a robust hybrid algorithm designed 
to maximize precision and reduce false positives, enabling safer and more effective decision-making in downstream perception modules.
To validate its effectiveness, we rigorously evaluate GSeg3D on the challenging SemanticKITTI benchmark, 
where it demonstrates strong performance for ground extraction in autonomous systems.

\section{Related Work}

Ground segmentation methods in autonomous driving and robotics typically fall into two categories: traditional techniques and deep 
learning-based approaches \cite{Gomes2023}.

Traditional methods primarily rely on geometric and statistical modeling of point cloud data. These methods typically utilize occupancy 
grids or elevation maps to spatially discretize the environment for efficient ground identification \cite{5650541,9558794}. 

Other traditional methods focus on modeling local surface characteristics using Gaussian process regression, plane fitting, 
and line extraction techniques. Himmelsbach et al. \cite{linefit} proposed a line extraction approach, achieving high precision 
up to 98.5\% in well-structured urban environments, but struggled in complex terrains. Gaussian Process Regression (GPR) methods 
\cite{GPR} have reported precision around 94.8\%, showing robust performance in urban and semi-structured settings. Similarly, 
CascadedSeg \cite{CascadedSeg} combined plane fitting and elevation thresholds, achieving precision of approximately 97\% in 
structured urban datasets like SemanticKITTI. Patchwork \cite{9466396} and its successor Patchwork++ \cite{9981561} are patch-based ground segmentation algorithms 
achieving high precision. Patchwork typically reaches 92–94\%, while Patchwork++ improves further to 93–95\%, due to enhanced local 
modeling and refined patch updates.

Region-growing approaches and clustering techniques further refine segmentation by exploiting spatial continuity. Zermas et al. 
\cite{GPF} employed a region-growing method based on planar characteristics, reporting precision of approximately 95\% but with 
significant drops in cluttered or vegetated environments. Methods such as R-GPF \cite{R-GPF}, a refinement of GPF using 
multi-resolution grids, enhanced the recall but faced challenges maintaining precision, dropping to 84\% in complex environments.

Advanced approaches using Conditional Random Field (CRF) and Markov Random Field (MRF) incorporate spatial coherence explicitly. 
Huang et al. \cite{9410344} demonstrated an MRF-based approach with precision values around 93\%, though the computational 
complexity was notably higher.

Deep learning approaches have gained traction due to their high flexibility and end-to-end training capabilities. Convolutional 
Neural Networks (CNN) initially adapted from object detection tasks have been successfully repurposed for ground segmentation in
point cloud data. PointNet \cite{qi2017pointnet} and PointNet++ \cite{qi2017pointnet++} offered unified frameworks achieving 
precision exceeding 96\% in various segmentation tasks, although they require extensive training data.

PointPillars \cite{lang2019pointpillars}, designed for real-time detection and segmentation in autonomous driving scenarios, 
reported precision rates near 97\% on SemanticKITTI datasets. SectorGSnet \cite{9691325} introduced an angular-sector-based 
neural network model specifically for ground segmentation, achieving precision up to 98\%, indicating strong potential for 
real-time autonomous driving applications.

Graph-based approaches such as GATA \cite{GraphBasedGS} and GndNet \cite{paigwar2020gndnet} employ attention mechanisms and 
graph neural networks. GndNet achieved precision scores around 96\%, demonstrating robustness in various urban datasets. 
Meanwhile, JCP \cite{rs13163239} combined CNN and point-based representations to handle complex terrains, achieving precision 
around 95\%, though computational overhead remains a challenge.

Overall, both traditional and learning-based methods achieve high precision in structured environments but frequently exhibit 
performance degradation in unstructured, cluttered, or vegetation-rich terrains. This emphasizes the continued importance of 
robust geometric and statistical approaches that ensure high precision for safety-critical autonomous systems.

\section{Proposed Solution}

\begin{figure}[tb]
\centering
\includegraphics[width=1.0\linewidth]{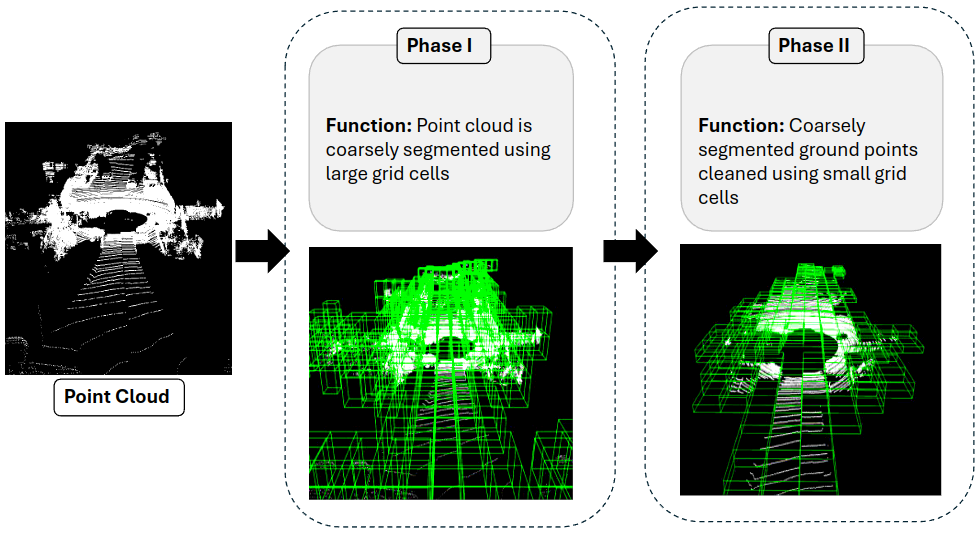}
\caption{Phase I uses grid cells with large height to capture tall structures as non-ground points. The coarse 
ground points from Phase I are subsequently processed in Phase II using grid cells with small height.}
\label{fig:phases}
\end{figure}

GSeg3D takes a single point cloud as input and outputs ground and non-ground points.
The segmentation is performed in two distinct phases (Fig.~\ref{fig:phases}).

\begin{figure}[tb]
\centering
\includegraphics[width=1.0\linewidth]{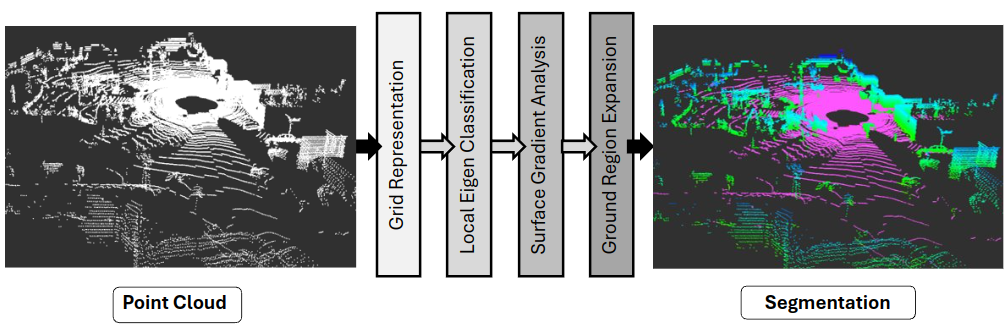}
\caption{Each phase consists of a four-step process. The figure displays the raw point cloud (in white), the detected ground points
 (highlighted in magenta), and the non-ground points, which are color-coded according to their height along the positive z-axis.}
\label{fig:pipeline}
\end{figure}

In the first phase, a larger grid cell height is used to capture as many non-ground structures as possible. 
This approach ensures that even elevated objects are quickly marked as non-ground, but it can occasionally 
misclassify some ground points, especially if the ground is partially covered by objects.

\begin{figure}[tb]
\centering
\includegraphics[width=0.8\linewidth,trim={0 0 0 1.5cm},clip]{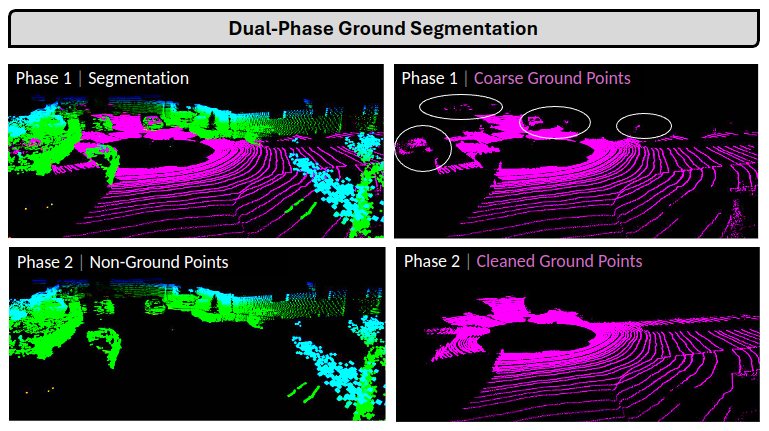}
\caption{Correction of over-segmentation of ground points based on dual-phase segmentation.}
\label{fig:dual_phase_segmentation}
\end{figure}

To address this, GSeg3D employs a second phase that uses a much smaller grid cell height. This phase 
refines the initial segmentation by re-examining points at finer vertical resolution, allowing the method 
to correct for false positives and false negatives from the first stage. Specifically, points initially 
misclassified are re-evaluated based on their local neighborhood and vertical structure, as illustrated 
in Fig.~\ref{fig:dual_phase_segmentation}.

This dual-phase design enables GSeg3D to achieve both high precision and high recall. By combining a 
coarse initial filtering with a targeted refinement, GSeg3D accurately distinguishes ground and 
non-ground points, which is critical for downstream tasks that require reliable terrain awareness.

Table~\ref{tab:phase_behavior} summarizes the differences between the two phases. The next sections 
describe each step in more detail.

\begin{table}[tb]
  \centering
  \caption{Processing steps in 1\textsuperscript{st} and 2\textsuperscript{nd} segmentation phase.}
  \label{tab:phase_behavior}
  \begin{tabular}{@{}p{3cm}p{2.2cm}p{3.1cm}@{}}
\toprule
  \textbf{Step} & \textbf{First Phase} & \textbf{Second Phase} \\
\midrule
  Grid Representation & Large cell height & Small cell height \\
  Local Eigen Classification & Identical & Identical \\
  Surface Gradient Analysis & Identical & Identical \\
  Ground Region Expansion & Neighbor expansion & Neighbor expansion and \newline additional point-wise check \\
\bottomrule
  \end{tabular}
\end{table}

\subsection{Grid Representation}

Each point in the input point cloud is assigned to a cell in a regularly spaced 3D grid. This discretization 
enables efficient modeling and detection of free, occupied, and ambiguous regions within the scene. The 
cell index for a given point is computed as:
\begin{align}
cell_{\{x,y,z\}} = \left\lfloor \frac{point_{\{x,y,z\}}}{cellsize_{\{x,y,z\}}} \right\rfloor
\end{align}
where $cellsize$ defines the spatial resolution of the grid in each dimension.

This grid-based representation forms the basis for subsequent segmentation steps. It enables fast 
neighbor queries and local geometric analysis, which are critical for distinguishing ground from 
non-ground points in large-scale point clouds.

\subsection{Local Eigen Classification}

\begin{figure}[tbp]
  \centering
  \includegraphics[width=0.8\linewidth]{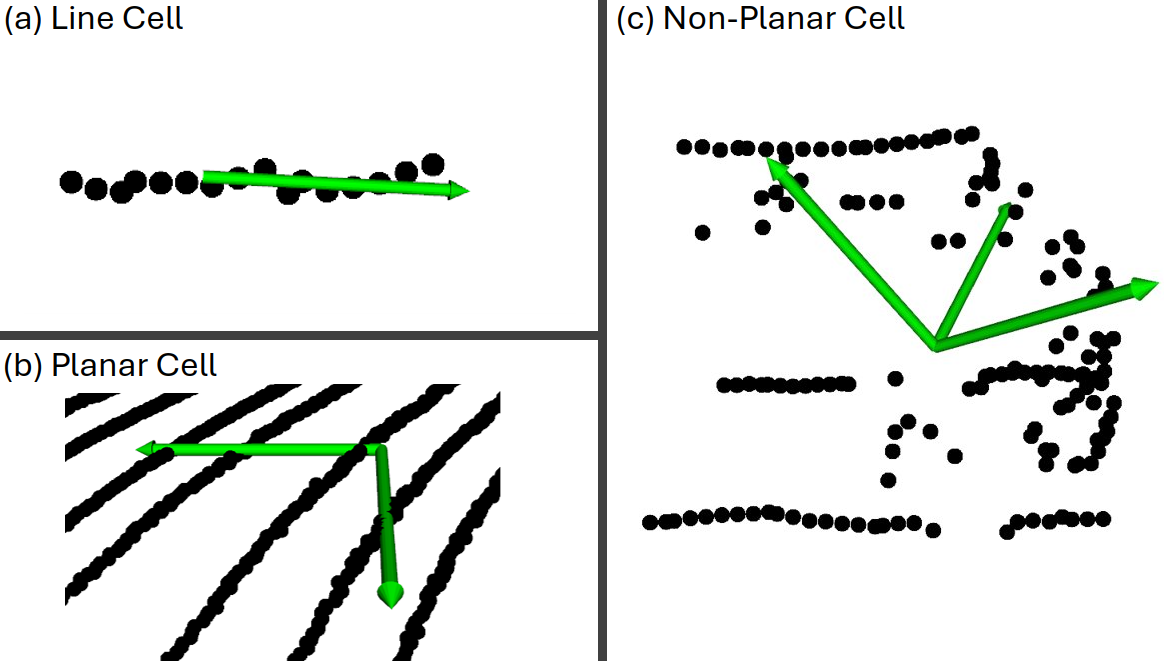}
  {\phantomsubcaption\label{fig:LineCell}}
  {\phantomsubcaption\label{fig:PlaneCell}}
  {\phantomsubcaption\label{fig:UnknownCell}}
  \caption{
    Points (black) assigned to a grid cell with dominant eigenvectors (green) are shown.
    The cell is classified as Line~(a), Planar~(b), or Non-Planar~(c) based on the local eigen classification.
  }
\end{figure}

We perform local eigen classification to characterize the spatial arrangement of points within each grid cell. For every cell, we first compute the \textit{covariance matrix} of the 3D coordinates of its points:  

\begin{equation}
\mathbf{C} = \frac{1}{N} \sum_{i=1}^{N} (\mathbf{p}_i - \bar{\mathbf{p}})(\mathbf{p}_i - \bar{\mathbf{p}})^{T}
\end{equation} where $\mathbf{p}_i \in \mathbb{R}^3$ denotes the coordinates of the $i$-th point, $\bar{\mathbf{p}}$ is the centroid of all points in the cell, and $N$ is the number of points.  

The eigenvalues $\lambda_1, \lambda_2$, and $\lambda_3$ of $\mathbf{C}$ describe how the points are distributed along the principal axes of variation:  

\begin{itemize}
    \item $\lambda_1 \gg \lambda_2, \lambda_3$: points are aligned mainly in one direction,
    \item $\lambda_1 \approx \lambda_2 \gg \lambda_3$: points lie predominantly on a plane,
    \item $\lambda_1 \approx \lambda_2 \approx \lambda_3$: points are scattered in all directions.
\end{itemize}
Let $\lambda_1 > \lambda_2 > \lambda_3$ denote the eigenvalues in decreasing order. We then compute the ratio:
\begin{align}
  \text{Ratio} = \tfrac{\lambda_1}{\lambda_1 + \lambda_2 + \lambda_3}
\end{align}

This ratio reveals whether the points are primarily 
aligned in a single direction (line), distributed along a surface (plane), or lack any clear 
geometric preference (non-planar).

\subsubsection{Line Cell}\label{sec:LineCell}
A cell is categorized as a Line Cell if $\lambda_1$ is much larger than $\lambda_2$ and $\lambda_3$, 
indicating strong alignment of points along a single direction. For such cells, we compute the angle 
between the principal eigenvector (corresponding to $\lambda_1$, shown in green in Fig.~\ref{fig:LineCell}) 
and the $z$-axis.

\begin{align}
\text{CellType} =
  \begin{smallcases}
    \text{Obstacle}         & \text{if } \text{angle} \geq 90^\circ - \text{Threshold} \\
    \text{Tentative Ground} & \text{if } \text{angle} < 90^\circ - \text{Threshold}
  \end{smallcases}\label{eq:Line:GroundThreshold}
\end{align}

Here, $Threshold$ is the maximum slope considered for ground points. Cells with an orientation below 
this slope threshold are provisionally marked as tentative ground, to be further verified in the region 
growth stage (see Section~\ref{sec:region_growth}).

\subsubsection{Planar Cell}
Cells with two dominant eigenvalues and a third, much smaller eigenvalue are classified as Planar Cells. 
This pattern indicates the points are distributed in a surface-like manner. However, because such 
distributions may arise from both ground surfaces and flat obstacle surfaces (e.g., roofs), Planar 
Cells are not directly labeled as ground without additional checks.

\subsubsection{Non-planar Cell}
If the eigenvalues are of similar magnitude, the cell is classified as Non-Planar. This suggests 
the points within the cell are scattered in space with no strong alignment, typical of clutter or 
irregular features.

\subsection{Surface Gradient Analysis}

\begin{figure}
\centering
\includegraphics[width=1.0\linewidth]{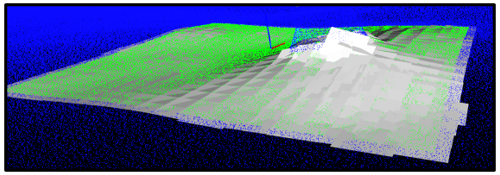}
\caption{To estimate the local surface slope, a plane model is fitted within each grid cell, allowing for a 
detailed and localized analysis of the terrain’s inclination.}
\label{fig:PlaneFit}
\end{figure}

For cells classified as planar in the local eigen classification, a robust plane fitting procedure is 
applied to accurately characterize the local surface geometry. We employ RANSAC-based plane fitting,
which separates the points in each planar cell into two categories: \textit{inliers}, which 
closely fit the estimated plane within a specified threshold, and \textit{outliers}, which 
lie farther away.

Once a plane is fitted, we calculate its normal vector and determine the slope angle with respect 
to the horizontal plane (the $xy$-plane), as shown in Fig.~\ref{fig:PlaneFit}. This slope angle 
serves as the principal criterion for further cell classification:

\begin{align}
\text{CellType} &=
\begin{smallcases}
\text{Tentative Ground}, & \text{if angle} \leq \text{Threshold} \\
\text{Non-Ground}, & \text{if angle} > \text{Threshold}
\end{smallcases} \label{eq:Planar:Ground}
\end{align}

Here, \textit{Threshold} denotes the maximum slope considered compatible with typical ground 
surfaces. Cells whose fitted plane slope is less than or equal to this threshold are labeled 
as \textit{tentative ground}; those with steeper slopes are classified as \textit{non-ground}. 
The tentative ground cells determined in this manner undergo further verification in the 
subsequent stages of the algorithm to ensure accurate segmentation.

This step is crucial for distinguishing between relatively flat ground regions and sloped surfaces 
that are unlikely to be traversable, improving the precision of the ground/non-ground segmentation.

\subsection{Ground Region Expansion}\label{sec:region_growth}

A well-known challenge in grid-based ground segmentation approaches is the dependency of region 
expansion on grid cell size. When the cell size is smaller than the gap between consecutive scan 
lines (as is common in sparse or non-uniform point clouds), traditional grid-based neighbor expansion 
fails. This leads to disconnected ground regions and poor recall, since expansion cannot bridge 
the gaps between scan lines. As a result, grid-based methods are often forced to use larger cell 
sizes at the expense of spatial precision and fine-grained segmentation.

GSeg3D addresses this limitation by adopting a KD-tree--based expansion strategy. All centroids 
of candidate ground cells are indexed using a KD-tree, enabling efficient and radius-based neighborhood 
search in three dimensions. This allows for region expansion even when cells are not directly adjacent 
in the grid, but are spatially proximate in the point cloud. As a result, much smaller grid cell sizes 
can be used without sacrificing connectivity or recall.

Combined with local eigen classification, which ensures that each cell’s geometric structure is accurately 
characterized, this approach enables high-precision and high-recall ground segmentation at finer 
spatial resolutions.

The selection of the initial seed cell for region expansion is critical: a poor seed can adversely 
affect segmentation quality. To ensure a reliable and robust start, GSeg3D injects synthetic 
points directly below the robot, based on the \texttt{robotRadius} and \texttt{distToGround} parameters. 
The cell corresponding to the position directly underneath the robot (i.e., the synthetic seed cell) 
is always used as the starting point for expansion, guaranteeing that the initial ground seed is valid, 
even in the presence of sensor gaps or occlusions. These synthetic points are removed prior to final 
metric calculation and output.

The expansion proceeds as follows:
\begin{itemize}
    \item The centroids of all tentative ground cells are indexed in a KD-tree for fast neighbor queries.
    \item Expansion is initialized from the seed cell directly below the robot.
    \item For each dequeued candidate cell, all neighboring centroids within a specified radius are queried via the KD-tree.
    \item In Phase II (refinement), only neighbors within a height difference threshold are considered to maintain consistency.
    \item Valid neighboring cells are queued for further expansion if they are also classified as tentative ground.
\end{itemize}

\begin{algorithm}[ht]
\caption{Ground Region Expansion}
\begin{algorithmic}[1]
\State \textbf{Config:} $r$: search radius, $\tau$: inlier threshold,  phase
\State \textbf{Input:} Seed queue $Q$, tentative ground centroids $C$, grid cells $\mathcal{G}$
\State \textbf{Output:} Segmented point sets $P_{\text{ground}}$, $P_{\text{non-ground}}$
\State Build KD-tree for centroids $C$
\While{$Q$ not empty}
    \State Dequeue cell $i$ from $Q$; mark $i$ as expanded
    \If{cell $i$ is empty}
        \State \textbf{continue}
    \EndIf
    \State // Perform radius search around centroid $c_i$ within $r$:
    \State $N = neighbors(c_i,r)$
    \For{each $j\in N$}
        \If{$j$ expanded or $j\in Q$}
            \State \textbf{continue}
        \EndIf
        \If{$phase = 2$ and $|c_{i,z} - c_{j,z}| > \tau$}
            \State \textbf{continue}
        \EndIf
        \State Mark cell $j$ as GROUND; enqueue $j$ to $Q$
    \EndFor
    \State Extract ground inliers and non-ground outliers in $i$
    \If{ground inliers are empty}
        \State \textbf{continue}
    \EndIf
    \State Classify sparsity for inliers and outliers
    \If{scores equal}
        \State Compute $z_i=height(i)$, gather $z_n$ of neighbors
        \If{neighbors empty \textbf{or} $|z_i - z_n| > \tau$}
            \State Mark as non-ground; \textbf{continue}
        \EndIf
        \If{non-ground cell exists below}
            \State Mark as non-ground; \textbf{continue}
        \EndIf
    \EndIf
    \State Add inliers to $P_{\text{ground}}$, outliers to $P_{\text{non-ground}}$
\EndWhile
\State \Return $P_{\text{ground}}, P_{\text{non-ground}}$
\end{algorithmic}
\end{algorithm}

After region expansion, each candidate cell undergoes further refinement using the following procedure, 
closely following the implemented logic:

\begin{enumerate}
\item \textbf{Extraction of Inliers and Outliers:}\
Each cell’s points are divided into \emph{ground inliers} (those close to the previously fitted plane) 
and \emph{non-ground inliers} (those further away), based on the indices identified during the Surface Gradient Analysis.

\item \textbf{Bounding Box Sparsity Analysis:}\\
For both sets of points (ground and non-ground inliers), a three-dimensional bounding box is computed and 
its sparsity is evaluated. The sparsity score, defined as the ratio of bounding box volume to the number 
of points, is categorized as ``Low,'' ``Medium,'' or ``High.'' If both inlier groups receive the same 
sparsity label, the bounding box analysis is considered ambiguous and additional checks are triggered.

\item \textbf{Ambiguity Resolution via Local Neighborhood:}\\
When bounding box scores are ambiguous, the algorithm compares the height of the centroid of the current 
cell’s ground inliers with those of neighboring ground cells. If no valid ground neighbors are found, or 
if the cell centroid is significantly elevated (by more than $0.3\,m$) above the lowest neighbor, the cell 
is rejected as ground to prevent mislabeling of elevated structures.

\item \textbf{Floating Cell Rejection:}\\
To avoid classifying floating structures as ground, the method checks for any occupied non-ground cell 
directly beneath the current cell in the vertical stack. If such a cell exists, the current cell is 
labeled as non-ground.

\item \textbf{Final Assignment:}\\
Only cells passing all previous checks are assigned as ground, while those failing any step are classified as non-ground.
\end{enumerate}

This multi-step refinement, combining geometric, density-based, and local contextual analysis, ensures 
that only spatially coherent and physically plausible regions are segmented as ground. The resulting 
segmentation achieves robust, high-precision ground extraction suitable for safety-critical applications.

The KD-tree driven iterative expansion thus enables robust, contiguous ground region extraction across 
the point cloud, even at high grid resolutions. This method overcomes the fundamental connectivity 
limitations of traditional grid-based expansion and is especially effective for safety-critical applications 
where accurate and dense ground segmentation is required.

\section{Experiments}

For comparison, we selected state-of-the-art ground segmentation algorithms as listed in 
Table~\ref{tab:results_main}, using their default or recommended parameters from the respective 
publications or open-source implementations. For all experiments with GSeg3D, we used a grid size of $(1.5,1.0,1.5)$m in Phase I and $(1.5,1.0,0.2)$m 
in Phase II. The ground inlier threshold was set to $0.125$m and the ground slope threshold to $30^\circ$. 
The following configuration was used:
\begin{verbatim}
distToGround: 1.723
robotRadius: 2.7
cellSizeX: 1.5
cellSizeY: 1.0
cellSizeZ: 1.5 (Phase I), 0.2 (Phase II)
slopeThresholdDegrees: 30.0
groundInlierThreshold: 0.125
centroidSearchRadius: 5.0
\end{verbatim}

\subsection{Dataset}
We evaluated all algorithms on the SemanticKITTI dataset~\cite{Geiger2013IJRR}, which consists of large-scale LiDAR point 
clouds acquired in urban environments. The data was collected using a car platform equipped with a Velodyne HDL-64E sensor mounted 
at a height of 1.72m. This dataset, recorded in 2013, provides high-resolution, annotated scans that serve as a standard benchmark 
for autonomous driving research.

\subsection{Performance Metrics}

To quantitatively evaluate segmentation performance, we use precision, recall, and F1-score as our primary metrics:
\begin{align}
  \text{Precision} &= \tfrac{\text{NTP}}{\text{NTP} + \text{NFP}} \\
  \text{Recall} &= \tfrac{\text{NTP}}{\text{NTP} + \text{NFN}} \\
  \text{F1-score} &= 2 \cdot \tfrac{\text{Precision} \cdot \text{Recall}}{\text{Precision} + \text{Recall}}
\end{align}
where
\begin{itemize}
\item \textbf{NTP}: Number of True Positives (correctly predicted ground points),
\item \textbf{NFP}: Number of False Positives (non-ground points incorrectly predicted as ground),
\item \textbf{NFN}: Number of False Negatives (ground points incorrectly predicted as non-ground).
\end{itemize}

To provide a robust and comprehensive comparison, we evaluate each algorithm across multiple distance thresholds 
(from 10m to 100m, in 10m increments) for each dataset sequence. For every distance threshold, we compute 
the standard metrics of precision, recall, and F1-score.

For clarity and conciseness in reporting, we aggregate the results for each sequence by taking the arithmetic 
mean and standard deviation of the metric values across all tested distances. Thus, the final result for each 
algorithm and sequence is presented as “mean $\pm$ std”, representing the average performance and its variability 
over the entire evaluation range.

This approach follows established practice in benchmarking and reflects both the typical performance and its 
stability with respect to varying perception ranges. All quantitative results in this paper are based on these metrics, 
with ground-truth labels provided by the SemanticKITTI dataset. The 
F1-score provides a harmonic mean between precision and recall, offering a balanced measure of segmentation performance.

\section{Results and Discussion}

The performance of eight ground segmentation algorithms was evaluated on sequences 00 to 10 of the SemanticKITTI dataset, with detailed 
quantitative comparisons presented in Tables~\ref{tab:results_main}, and \ref{tab:average_results}. 
Metrics reported include precision (Pr.), recall (Rc.), and F1-score, each accompanied by standard deviations (std).

\begin{table*}[ht]
\centering
\scriptsize
\caption{Comparison of ground segmentation methods on SemanticKITTI (sequences 00--10). Results are 
reported as mean $\pm$ std (\%).}
\label{tab:results_main}
\begin{tabular}{lccccccccccccc}
\toprule
\multirow{2}{*}{\textbf{Method}} & 
\multicolumn{3}{c}{\textbf{00}} & 
\multicolumn{3}{c}{\textbf{01}} & 
\multicolumn{3}{c}{\textbf{02}} & 
\multicolumn{3}{c}{\textbf{03}} \\
\cmidrule(lr){2-4} \cmidrule(lr){5-7} \cmidrule(lr){8-10} \cmidrule(lr){11-13}
 & Pr. & Rc. & F1 & Pr. & Rc. & F1 & Pr. & Rc. & F1 & Pr. & Rc. & F1 &  \\
\midrule
CGSeg & 91.6$\pm$9.1 & 71.0$\pm$10.6 & 80.0 & \textbf{97.2}$\pm$2.9 & 75.7$\pm$10.5 & 85.1 & 91.8$\pm$7.7 & 70.0$\pm$7.7 & 79.4 & 92.4$\pm$8.0 & 69.0$\pm$8.1 & 79.0  \\
GPF & 95.1$\pm$4.9 & 80.9$\pm$25.5 & 87.4 & 93.9$\pm$6.4 & 42.6$\pm$23.5 & 55.7 & 96.4$\pm$3.5 & 79.0$\pm$18.5 & 86.8 & \textbf{97.5}$\pm$\textbf{2.8} & 78.7$\pm$13.5 & 87.0  \\
Linefit & \textbf{98.2}$\pm$\textbf{1.0} & 81.1$\pm$10.0 & 88.8 & \textbf{99.0}$\pm$\textbf{1.4} & 74.6$\pm$9.6 & 84.9 & \textbf{98.0}$\pm$\textbf{1.2} & 82.5$\pm$5.9 & 89.5 & \textbf{95.8}$\pm$4.1 & 79.3$\pm$8.4 & 86.6  \\
Patchwork & 92.2$\pm$3.5 & 93.3$\pm$3.9 & 92.7 & 95.6$\pm$3.5 & 89.8$\pm$5.2 & 92.6 & 92.1$\pm$3.9 & 93.2$\pm$\textbf{3.8} & \textbf{92.6} & 89.4$\pm$7.1 & \textbf{95.3}$\pm$\textbf{2.7} & 92.2  \\
Patchwork++ & 93.8$\pm$3.9 & \textbf{93.9}$\pm$\textbf{3.3} & \textbf{93.9} & 89.9$\pm$6.4 & \textbf{97.4}$\pm$\textbf{2.8} & \textbf{93.5} & 93.8$\pm$3.9 & \textbf{93.6}$\pm$3.9 & \textbf{93.7} & 94.7$\pm$\textbf{2.9} & 91.0$\pm$6.6 & \textbf{92.8} \\
R\_GPF & 56.5$\pm$11.4 & \textbf{96.2}$\pm$\textbf{1.7} & 71.2 & 88.0$\pm$6.4 & 90.6$\pm$4.3 & 89.3 & 65.6$\pm$13.1 & \textbf{96.4}$\pm$\textbf{1.7} & 78.1 & 72.0$\pm$10.1 & \textbf{96.8}$\pm$\textbf{1.3} & 82.6 \\
Ransac & 87.9$\pm$15.6 & 93.7$\pm$12.4 & 90.7 & 96.9$\pm$\textbf{2.2} & \textbf{96.3}$\pm$\textbf{3.5} & \textbf{96.6} & 90.9$\pm$13.8 & 90.9$\pm$8.9 & 90.9 & 92.8$\pm$7.1 & 90.3$\pm$8.8 & 91.5  \\
\specialrule{1.2pt}{1pt}{1pt} 
GSeg3D & \textbf{97.1}$\pm$\textbf{2.1} & 89.9$\pm$5.7 & \textbf{93.4} & \textbf{97.2}$\pm$3.7 & 85.5$\pm$7.9 & 90.9 & \textbf{97.0}$\pm$\textbf{2.2} & 90.7$\pm$4.2 & \textbf{93.7} & 94.1$\pm$6.2 & 92.0$\pm$4.3 & \textbf{93.0} \\
\bottomrule
\end{tabular}
\vskip5pt
%
\begin{tabular}{lccccccccccccc}
\toprule
\multirow{2}{*}{\textbf{Method}} & \multicolumn{3}{c}{\textbf{04}} & \multicolumn{3}{c}{\textbf{05}} & \multicolumn{3}{c}{\textbf{06}} & \multicolumn{3}{c}{\textbf{07}} \\
\cmidrule(lr){2-4} \cmidrule(lr){5-7} \cmidrule(lr){8-10} \cmidrule(lr){11-13} 
 & Pr. & Rc. & F1 & Pr. & Rc. & F1 & Pr. & Rc. & F1 & Pr. & Rc. & F1 & \\
\midrule
CGSeg & 98.2$\pm$2.5 & 73.7$\pm$4.2 & 84.2 & 86.1$\pm$12.9 & 68.4$\pm$12.4 & 76.3 & 98.2$\pm$2.3 & 75.9$\pm$4.6 & 85.6 & 89.6$\pm$10.7 & 72.4$\pm$11.1 & 80.1 \\
GPF & 95.9$\pm$4.6 & 76.1$\pm$25.9 & 84.8 & 93.4$\pm$4.1 & 88.8$\pm$9.1 & 91.0 & 97.4$\pm$2.2 & 78.9$\pm$17.0 & 87.1 & 94.3$\pm$3.6 & 89.5$\pm$17.0 & 91.8 \\
Linefit & \textbf{99.5}$\pm$\textbf{0.6} & 78.2$\pm$4.7 & 87.5 & \textbf{96.7}$\pm$\textbf{1.6} & 83.2$\pm$6.8 & 89.3 & \textbf{99.3}$\pm$\textbf{0.4} & 80.5$\pm$3.5 & 88.8 & \textbf{97.2}$\pm$\textbf{1.3} & 87.6$\pm$5.6 & 92.0 \\
Patchwork & 97.4$\pm$1.4 & 91.5$\pm$2.6 & \textbf{94.4} & 89.3$\pm$5.0 & \textbf{94.2}$\pm$\textbf{2.8} & 91.6 & 96.8$\pm$\textbf{1.3} & 96.2$\pm$\textbf{1.3} & \textbf{96.5} & 90.2$\pm$4.2 & \textbf{95.5}$\pm$\textbf{2.5} & 92.7\\
Patchwork++ & 90.2$\pm$3.5 & \textbf{97.9}$\pm$\textbf{1.3} & 93.9 & \textbf{94.9}$\pm$2.8 & 91.1$\pm$4.4 & \textbf{93.0} & 96.4$\pm$1.8 & \textbf{97.7}$\pm$\textbf{1.0} & \textbf{97.1} & \textbf{96.4}$\pm$2.6 & 91.9$\pm$3.4 & \textbf{94.1} \\
R\_GPF & 83.0$\pm$9.0 & 93.4$\pm$1.6 & 87.9 & 57.7$\pm$15.0 & \textbf{96.2}$\pm$\textbf{1.3} & 72.1 & 82.2$\pm$7.2 & 95.6$\pm$\textbf{1.3} & 88.4 & 55.3$\pm$13.3 & \textbf{97.1}$\pm$\textbf{1.2} & 70.4  \\
Ransac & 96.1$\pm$2.1 & \textbf{96.9}$\pm$\textbf{1.4} & \textbf{96.5} & 85.7$\pm$13.5 & 93.8$\pm$7.5 & 89.6 & 95.5$\pm$2.6 & \textbf{97.3}$\pm$1.8 & 96.4 & 84.4$\pm$18.0 & 94.4$\pm$11.7 & 89.1 \\
\specialrule{1.2pt}{1pt}{1pt} 
GSeg3D & \textbf{99.3}$\pm$\textbf{1.2} & 89.5$\pm$2.8 & 94.1 & \textbf{94.9}$\pm$\textbf{2.5} & 89.2$\pm$4.3 & \textbf{92.0} & \textbf{98.5}$\pm$\textbf{1.3} & 89.4$\pm$3.8 & 93.6 & 96.1$\pm$\textbf{2.2} & 90.9$\pm$3.4 & \textbf{93.4} \\
\bottomrule
\end{tabular}
\vskip5pt
%
\begin{tabular}{lcccccccccc}
\toprule
\multirow{2}{*}{\textbf{Method}} & \multicolumn{3}{c}{\textbf{08}} & \multicolumn{3}{c}{\textbf{09}} & \multicolumn{3}{c}{\textbf{10}} \\
\cmidrule(lr){2-4} \cmidrule(lr){5-7} \cmidrule(lr){8-10} 
 & Pr. & Rc. & F1 & Pr. & Rc. & F1 & Pr. & Rc. & F1 & \\
\midrule
CGSeg & 90.3$\pm$13.9 & 67.1$\pm$13.5 & 77.0 & 94.8$\pm$5.9 & 71.0$\pm$8.0 & 81.2 & 82.7$\pm$20.8 & 59.2$\pm$17.5 & 69.0 \\
GPF & 96.9$\pm$3.7 & 76.0$\pm$21.3 & 85.1 & 96.2$\pm$5.7 & 69.0$\pm$25.2 & 80.2 & 91.7$\pm$7.2 & 66.5$\pm$23.2 & 76.9 \\
Linefit & \textbf{98.7}$\pm$\textbf{1.0} & 77.2$\pm$9.1 & 86.5 & \textbf{98.0}$\pm$\textbf{1.3} & 78.8$\pm$8.7 & 87.3 & \textbf{97.0}$\pm$\textbf{1.4} & 76.7$\pm$10.9 & 85.6 \\
Patchwork & 93.2$\pm$4.6 & 92.5$\pm$\textbf{4.6} & \textbf{92.8} & 91.5$\pm$4.0 & 92.2$\pm$4.2 & 91.8 & 88.4$\pm$5.0 & \textbf{89.5}$\pm$\textbf{7.3} & 88.9 \\
Patchwork++ & 93.3$\pm$4.1 & \textbf{94.5}$\pm$\textbf{4.6} & \textbf{93.9} & 92.1$\pm$4.4 & \textbf{93.8}$\pm$\textbf{3.7} & \textbf{92.9} & 91.0$\pm$5.1 & 87.5$\pm$8.9 & \textbf{89.2} \\
R\_GPF & 64.8$\pm$18.8 & \textbf{95.2}$\pm$\textbf{2.5} & 77.1 & 68.8$\pm$12.7 & \textbf{95.2}$\pm$\textbf{3.0} & 79.8 & 53.0$\pm$19.2 & \textbf{95.4}$\pm$\textbf{2.7} & 68.1  \\
Ransac & 85.8$\pm$23.8 & 88.7$\pm$18.6 & 87.2 & 93.5$\pm$8.0 & 89.9$\pm$6.1 & 91.7 & 72.4$\pm$28.2 & 82.6$\pm$19.8 & 77.2  \\
\specialrule{1.2pt}{1pt}{1pt} 
GSeg3D & \textbf{98.0}$\pm$\textbf{2.0} & 86.9$\pm$7.4 & 92.1 & \textbf{96.8}$\pm$\textbf{2.5} & 88.8$\pm$4.8 & \textbf{92.6} & \textbf{94.2}$\pm$\textbf{3.4} & 85.0$\pm$8.3 & \textbf{89.4} \\
\bottomrule
\end{tabular}
\end{table*}

\subsection{Urban Environments (Sequences 00, 02, 05, 06, 07, 09)}
Urban sequences typically involve structured roadways, sidewalks, and clutter from parked cars and street furniture. In these scenarios, \textit{Linefit} 
consistently demonstrated the highest precision (often exceeding 98\%) due to its linear-model assumptions aligning well with structured roads. However, 
\textit{GSeg3D}, utilizing a localized grid-based spatial reasoning approach, closely matched this high precision (above 97\%) while offering improved 
adaptability to complex urban geometries and irregularities such as curbs, ramps, and potholes.

\textit{Patchwork} and \textit{Patchwork++} maintained stable and high precision (approximately 93-96\%), benefitting from their iterative patch-based 
refinement suited to urban contexts.

Regarding recall in urban environments, \textit{Patchwork++} and \textit{Patchwork} consistently achieved high recall values (typically above 90\%), 
benefiting from their localized segmentation strategies that effectively capture ground points even in cluttered scenes. \textit{GSeg3D}, while 
maintaining robust recall performance (generally between 85\% and 90\%), exhibited slightly lower values compared to the Patchwork-based methods. 
Nevertheless, its recall remains well-balanced and sufficiently high for reliable ground segmentation in urban scenarios. Conversely, methods such as 
\textit{CGSeg} and \textit{GPF}, which exhibited significantly lower recall (below 80\% in some urban sequences), struggled in densely cluttered areas or 
scenes with abrupt terrain transitions.

In urban contexts, the F1-score revealed \textit{GSeg3D} matched closely \textit{Patchwork++} as a highly balanced solution (93\%-95\%). However, \textit{GSeg3D's} 
high recall is due to consistently high precision values (above 97\%), whereas \textit{Patchwork++} has higher recalls (above 91\%).

\subsection{Suburban and Highway Environments (Sequences 01, 03, 04)}
Suburban and highway sequences are characterized by open roads, moderate clutter, consistent, and predictable terrain. Here, methods like \textit{Linefit} 
again excelled in precision due to the predominantly linear road geometry (precision $\sim$99\%). However, \textit{GSeg3D} still showed comparable performance 
(97-99\%), leveraging its grid-based segmentation to maintain robust adaptability around road edges, lane transitions, and minor roadside variations.

Recall performance in these environments strongly favored algorithms with localized approaches such as \textit{Patchwork++}, achieving recall values above 
95\%. \textit{GSeg3D} similarly showed consistently strong recall (around 89-91\%), demonstrating balanced detection capability essential for highway automation,
where consistent ground segmentation ensures stable vehicle control.

In terms of F1-score, \textit{Patchwork++} (around 94\%) and \textit{GSeg3D} (92-94\%) were particularly robust, balancing precision and recall effectively, 
making them suitable candidates for suburban autonomous driving scenarios.

\subsection{Complex and Vegetation-Rich Environments (Sequences 08, 10)}
Complex terrains featuring dense vegetation, uneven ground, and irregular structures presented the greatest segmentation challenge. Here, global fitting methods 
such as \textit{Linefit} and \textit{Ransac} faced higher variability and performance degradation, particularly noticeable in their recall (below 80\%).

In contrast, the localized, grid-based strategy of \textit{GSeg3D} allowed it to adapt effectively, maintaining high precision (above 94\%) and moderate recall 
(85-87\%) despite increased environmental complexity. \textit{R\_GPF}, characterized by high recall (above 95\%) yet lower precision (below 70\%), proved valuable 
for exploratory tasks where capturing ground points takes precedence over precision.

\textit{Patchwork++} maintained balanced precision and recall, highlighting its suitability for general-purpose deployment across mixed or uncertain terrains. 
The stability in the F1-score (89-93\%) demonstrated by \textit{GSeg3D} coupled with high precision (above 94\%) across these challenging sequences further confirmed 
its robustness for applications such as off-road robotics and outdoor mobile mapping.

\begin{table}[ht]
\centering
\scriptsize
\caption{Average performance (\%) across all sequences and Runtime (Rt.) in milliseconds.}
\label{tab:average_results}
\begin{tabular}{lccccccc}
\toprule
\textbf{Method} & Pr. & Rc. & F1 & Rt. & Pr. Std & Rc. Std & Rt. Std\\
\midrule
CGSeg & 92.1 & 70.3 & 79.7 & 59.1 & 8.8 & 9.8 & 11.0 \\
GPF & 95.3 & 75.1 & 83.1 & \textbf{11.4} & 4.4 & 20.0 & \textbf{0.7}\\
Linefit & \textbf{97.9} & 80.0 & 87.9 & \textbf{6.0} & \textbf{1.4} & 7.6 & \textbf{0.4} \\
Patchwork & 92.4 & 93.0 & 92.6 & 15.7 & 4.0 & \textbf{3.7} & 1.0 \\
Patchwork++ & 93.3 & \textbf{93.7} & \textbf{93.5} & 13.2 & 3.8 & 4.0 & 1.6 \\
R\_GPF & 67.9 & \textbf{95.3} & 78.6 & 17.4 & 12.4 & \textbf{2.0} & 1.1 \\
Ransac & 89.3 & 92.3 & 90.7 & 53.6 & 12.3 & 9.1 & 10.3 \\
\specialrule{1.2pt}{1pt}{1pt} 
GSeg3D & \textbf{96.6} & 89.4 & \textbf{92.8} & 48.1 &  \textbf{2.7} & 5.1 & 7.3 \\
\bottomrule
\end{tabular}
\end{table}

\subsection{Overall Performance Analysis}

Table~\ref{tab:average_results} summarizes the average performance metrics across all SemanticKITTI sequences.

\textbf{Precision:} \textit{Linefit} achieved the highest average precision (97.9\%) with the lowest variability (1.4\%), making it highly reliable in structured environments. \textit{GSeg3D} followed closely with high precision (96.6\%) and low variability (2.7\%), offering consistently strong performance across diverse and complex scenarios.

\textbf{Recall:} \textit{R\_GPF} achieved the highest recall (95.3\%) but suffered from low precision. In contrast, \textit{Patchwork++}, \textit{Patchwork}, and \textit{GSeg3D} offered balanced recall performance (93.7\%, 93.0\%, and 89.4\%, respectively), with \textit{GSeg3D} maintaining superior precision.

\textbf{F1-score:} \textit{Patchwork++} (93.5\%) and \textit{GSeg3D} (92.8\%) achieved the highest F1-scores, demonstrating balanced accuracy and completeness.

\textbf{Runtime:} \textit{Linefit} was fastest (6.0~ms), followed by \textit{GPF} and \textit{Patchwork++}. While slower (48.1~ms), \textit{GSeg3D} provides a solid trade-off between efficiency and segmentation quality.

\textbf{Stability:} \textit{GSeg3D} exhibited stable precision (2.7\% std dev), second only to \textit{Linefit}, reinforcing its predictable performance.

In summary, \textit{GSeg3D} offers an effective balance of high precision, reliable recall, stable output, and reasonable runtime, making it well-suited for real-world, safety-critical applications.

\subsection{Failure Case Analysis and Runtime Variability}

Our evaluation also highlights the limitations and variability observed across different environments. 
In vegetation-rich or highly cluttered scenarios (e.g., Sequences 08 and 10), global fitting methods such as \textit{Linefit} and \textit{RANSAC} exhibited reduced recall, 
while recall-focused methods like \textit{R\_GPF} maintained high completeness at the cost of precision. 
\textit{GSeg3D}, although robust, also showed moderate recall degradation in these conditions, reflecting the inherent difficulty of distinguishing ground points from 
low-lying vegetation or occluded regions. 

Future work will involve combining geometric reasoning with semantic cues (e.g., vegetation class segmentation) to improve the discrimination 
between true ground and vegetation, thereby enhancing robustness in such environments.

Regarding runtime, results in Table~\ref{tab:average_results} demonstrate that \textit{Linefit} and \textit{GPF} are the fastest, with very low variability (std $<1$ ms). 
In contrast, iterative methods such as \textit{Patchwork++} and the proposed \textit{GSeg3D} exhibit higher runtimes (13--48 ms) with greater variability 
due to their grid-based reasoning and refinement stages. 
Nevertheless, \textit{GSeg3D} achieves a favorable balance: while not the fastest, its runtime remains practical for real-time applications, 
and its higher variability is compensated by consistently strong segmentation quality.

\textbf{Application Suitability:} Summary analysis suggests:
\begin{itemize}
    \item \textbf{GSeg3D:} Best for safety-critical applications (autonomous driving, robotics) due to consistently high precision, good recall, and low variability.
    \item \textbf{Patchwork++ / Patchwork:} General-purpose methods with balanced precision and recall, suitable for urban mapping and general navigation.
    \item \textbf{Linefit:} Ideal for structured environments needing very high precision with minimal variability (e.g., infrastructure inspection).
    \item \textbf{R\_GPF:} Suited for exploratory or recall-focused tasks, where completeness matters more than precision.
    \item \textbf{Ransac:} Suitable for moderate-complexity scenarios and prototyping, balancing performance with higher variability.
    \item \textbf{GPF:} Applicable to static or controlled environments where occasional inaccuracies are acceptable.
    \item \textbf{CGSeg:} Useful for precision-focused tasks like boundary refinement or post-processing where lower recall is acceptable.
\end{itemize}

\section{Conclusion}

Overall, the results demonstrate that \textit{GSeg3D} offers near-top precision with notably low standard deviation, ensuring stable and predictable performance. 
Compared to \textit{Linefit}, which achieves slightly higher precision, \textit{GSeg3D} maintains significantly better recall while preserving high precision, 
resulting in a more balanced segmentation performance. 

Unlike methods biased toward either precision or recall, \textit{GSeg3D} consistently combines precise ground classification with reliable recall across diverse environments. 
This combination of high precision, good recall, low variability, and reasonable runtime makes \textit{GSeg3D} particularly suitable for safety-critical applications in 
autonomous driving and robotics, where both reliability and consistency are essential.

Nonetheless, challenges remain in vegetation-rich or occluded environments, where recall performance degrades due to the difficulty of separating ground points from 
low vegetation or partially visible surfaces. Future work will focus on addressing these cases by combining geometric reasoning with semantic segmentation cues 
(e.g., vegetation class detection), enabling more robust ground segmentation across highly complex outdoor scenarios.

\section{Acknowledgment}
We greatly acknowledge the open source Ground Segmentation Benchmark \cite{9466396,9981561} which we 
used for benchmarking. PCL \cite{PCL} and Nanoflann \cite{NANOFLANN} were used for 
processing the point clouds.

\bibliography{paper}

@ARTICLE{1,
  author={Anand, Bhaskar and Senapati, Mrinal and Barsaiyan, Vivek and Rajalakshmi, Pachamuthu},
  journal={IEEE Transactions on Instrumentation and Measurement}, 
  title={{LiDAR-INS}/{GNSS}-Based Real-Time Ground Removal, Segmentation, and Georeferencing Framework for Smart Transportation}, 
  year={2021},
  volume={70},
  number={},
  pages={1-11},
  doi={10.1109/TIM.2021.3117661}}

@INPROCEEDINGS{2,
  author={Arora, Mehul and Wiesmann, Louis and Chen, Xieyuanli and Stachniss, Cyrill},
  booktitle={2021 European Conference on Mobile Robots (ECMR)}, 
  title={Mapping the Static Parts of Dynamic Scenes from 3D {LiDAR} Point Clouds Exploiting Ground Segmentation}, 
  year={2021},
  volume={},
  number={},
  pages={1-6},
  doi={10.1109/ECMR50962.2021.9568799}}

@Article{Gomes2023,
AUTHOR = {Gomes, Tiago and Matias, Diogo and Campos, André and Cunha, Luís and Roriz, Ricardo},
TITLE = {A Survey on Ground Segmentation Methods for Automotive LiDAR Sensors},
JOURNAL = {Sensors},
VOLUME = {23},
YEAR = {2023},
NUMBER = {2},
ARTICLE-NUMBER = {601},
URL = {https://www.mdpi.com/1424-8220/23/2/601},
PubMedID = {36679414},
ISSN = {1424-8220},
DOI = {10.3390/s23020601}}

@ARTICLE{Jimenez2021,
  author={Jiménez, Víctor and Godoy, Jorge and Artuñedo, Antonio and Villagra, Jorge},
  journal={IEEE Access}, 
  title={Ground Segmentation Algorithm for Sloped Terrain and Sparse {LiDAR} Point Cloud}, 
  year={2021},
  volume={9},
  number={},
  pages={132914-132927},
  doi={10.1109/ACCESS.2021.3115664}}

@INPROCEEDINGS{CascadedSeg,
  author={Narksri, Patiphon and Takeuchi, Eijiro and Ninomiya, Yoshiki and Morales, Yoichi and Akai, Naoki and Kawaguchi, Nobuo},
  booktitle={2018 21st International Conference on Intelligent Transportation Systems (ITSC)}, 
  title={A Slope-robust Cascaded Ground Segmentation in 3D Point Cloud for Autonomous Vehicles}, 
  year={2018},
  volume={},
  number={},
  pages={497-504},
  doi={10.1109/ITSC.2018.8569534}}

@article{Arora2023,
title = {Static map generation from 3D {LiDAR} point clouds exploiting ground segmentation},
journal = {Robotics and Autonomous Systems},
volume = {159},
pages = {104287},
year = {2023},
issn = {0921-8890},
doi = {https://doi.org/10.1016/j.robot.2022.104287},
url = {https://www.sciencedirect.com/science/article/pii/S0921889022001762},
author = {Mehul Arora and Louis Wiesmann and Xieyuanli Chen and Cyrill Stachniss}
}

@article{Firkat2023,
title = {{FGSeg}: Field-ground segmentation for agricultural robot based on {LiDAR}},
journal = {Computers and Electronics in Agriculture},
volume = {211},
pages = {107965},
year = {2023},
issn = {0168-1699},
doi = {https://doi.org/10.1016/j.compag.2023.107965},
url = {https://www.sciencedirect.com/science/article/pii/S0168169923003538},
author = {Eksan Firkat and Fan An and Bei Peng and Jinlai Zhang and Tayir Mijit and Arzigul Ahat and Jihong Zhu and Askar Hamdulla}
}

@article{Geiger2013IJRR,
  author = {Andreas Geiger and Philip Lenz and Christoph Stiller and Raquel Urtasun},
  title = {Vision meets Robotics: The {KITTI} Dataset},
  journal = {International Journal of Robotics Research (IJRR)},
  year = {2013}
}

@inproceedings{qi2017pointnet,
  title={Pointnet: Deep learning on point sets for 3d classification and segmentation},
  author={Qi, Charles R and Su, Hao and Mo, Kaichun and Guibas, Leonidas J},
  booktitle={Proceedings of the IEEE conference on computer vision and pattern recognition},
  pages={652--660},
  year={2017}
}

@article{qi2017pointnet++,
  title={Pointnet++: Deep hierarchical feature learning on point sets in a metric space},
  author={Qi, Charles Ruizhongtai and Yi, Li and Su, Hao and Guibas, Leonidas J},
  journal={Advances in neural information processing systems},
  volume={30},
  year={2017}
}

@inproceedings{paigwar2020gndnet,
  title={{GndNet}: Fast ground plane estimation and point cloud segmentation for autonomous vehicles},
  author={Paigwar, Anshul and Erkent, {\"O}zg{\"u}r and Sierra-Gonzalez, David and Laugier, Christian},
  booktitle={2020 IEEE/RSJ International Conference on Intelligent Robots and Systems (IROS)},
  pages={2150--2156},
  year={2020},
  organization={IEEE}
}

@inproceedings{lang2019pointpillars,
  title={Pointpillars: Fast encoders for object detection from point clouds},
  author={Lang, Alex H and Vora, Sourabh and Caesar, Holger and Zhou, Lubing and Yang, Jiong and Beijbom, Oscar},
  booktitle={Proceedings of the IEEE/CVF conference on computer vision and pattern recognition},
  pages={12697--12705},
  year={2019}
}

@INPROCEEDINGS{5650541,
  author={Douillard, B. and Underwood, J. and Melkumyan, N. and Singh, S. and Vasudevan, S. and Brunner, C. and Quadros, A.},
  booktitle={2010 IEEE/RSJ International Conference on Intelligent Robots and Systems}, 
  title={Hybrid elevation maps: 3D surface models for segmentation}, 
  year={2010},
  volume={},
  number={},
  pages={1532-1538},
  doi={10.1109/IROS.2010.5650541}}

@ARTICLE{9558794,
  author={Anand, Bhaskar and Senapati, Mrinal and Barsaiyan, Vivek and Rajalakshmi, Pachamuthu},
  journal={IEEE Transactions on Instrumentation and Measurement}, 
  title={LiDAR-INS/GNSS-Based Real-Time Ground Removal, Segmentation, and Georeferencing Framework for Smart Transportation}, 
  year={2021},
  volume={70},
  number={},
  pages={1-11},
  doi={10.1109/TIM.2021.3117661}}

@ARTICLE{9410344,
  author={Huang, Weixin and Liang, Huawei and Lin, Linglong and Wang, Zhiling and Wang, Shaobo and Yu, Biao and Niu, Runxin},
  journal={IEEE Transactions on Intelligent Transportation Systems}, 
  title={A Fast Point Cloud Ground Segmentation Approach Based on Coarse-To-Fine Markov Random Field}, 
  year={2022},
  volume={23},
  number={7},
  pages={7841-7854},
  doi={10.1109/TITS.2021.3073151}}

@Article{rs13163239,
  AUTHOR = {Shen, Zhihao and Liang, Huawei and Lin, Linglong and Wang, Zhiling and Huang, Weixin and Yu, Jie},
  TITLE = {Fast Ground Segmentation for 3D LiDAR Point Cloud Based on Jump-Convolution-Process},
  JOURNAL = {Remote Sensing},
  VOLUME = {13},
  YEAR = {2021},
  NUMBER = {16},
  ARTICLE-NUMBER = {3239},
  URL = {https://www.mdpi.com/2072-4292/13/16/3239},
  ISSN = {2072-4292},
  DOI = {10.3390/rs13163239}
  }

@ARTICLE{9691325,
  author={He, Dong and Abid, Furqan and Kim, Young-Min and Kim, Jong-Hwan},
  journal={IEEE Access}, 
  title={SectorGSnet: Sector Learning for Efficient Ground Segmentation of Outdoor LiDAR Point Clouds}, 
  year={2022},
  volume={10},
  number={},
  pages={11938-11946},
  doi={10.1109/ACCESS.2022.3146317}}

@ARTICLE{9466396,
  author={Lim, Hyungtae and Oh, Minho and Myung, Hyun},
  journal={IEEE Robotics and Automation Letters}, 
  title={Patchwork: Concentric Zone-Based Region-Wise Ground Segmentation With Ground Likelihood Estimation Using a 3D LiDAR Sensor}, 
  year={2021},
  volume={6},
  number={4},
  pages={6458-6465},
  doi={10.1109/LRA.2021.3093009}}

@INPROCEEDINGS{9981561,
  author={Lee, Seungjae and Lim, Hyungtae and Myung, Hyun},
  booktitle={2022 IEEE/RSJ International Conference on Intelligent Robots and Systems (IROS)}, 
  title={Patchwork++: Fast and Robust Ground Segmentation Solving Partial Under-Segmentation Using 3D Point Cloud}, 
  year={2022},
  volume={},
  number={},
  pages={13276-13283},
  doi={10.1109/IROS47612.2022.9981561}}

@INPROCEEDINGS{linefit,
  author={Himmelsbach, M. and Hundelshausen, Felix v. and Wuensche, H.-J.},
  booktitle={2010 IEEE Intelligent Vehicles Symposium}, 
  title={Fast segmentation of 3D point clouds for ground vehicles}, 
  year={2010},
  volume={},
  number={},
  pages={560-565},
  doi={10.1109/IVS.2010.5548059}}

@INPROCEEDINGS{GPF,
  author={Zermas, Dimitris and Izzat, Izzat and Papanikolopoulos, Nikolaos},
  booktitle={2017 IEEE International Conference on Robotics and Automation (ICRA)}, 
  title={Fast segmentation of 3D point clouds: A paradigm on LiDAR data for autonomous vehicle applications}, 
  year={2017},
  volume={},
  number={},
  pages={5067-5073},
  doi={10.1109/ICRA.2017.7989591}}

@incollection{RANSAC,
title = {Random Sample Consensus: A Paradigm for Model Fitting with Applications to Image Analysis and Automated Cartography},
editor = {Martin A. Fischler and Oscar Firschein},
booktitle = {Readings in Computer Vision},
publisher = {Morgan Kaufmann},
address = {San Francisco (CA)},
pages = {726-740},
year = {1987},
isbn = {978-0-08-051581-6},
doi = {https://doi.org/10.1016/B978-0-08-051581-6.50070-2},
url = {https://www.sciencedirect.com/science/article/pii/B9780080515816500702},
author = {Martin A. Fischler and Robert C. Bolles}
}

@ARTICLE{R-GPF,
  author={Lim, Hyungtae and Hwang, Sungwon and Myung, Hyun},
  journal={IEEE Robotics and Automation Letters}, 
  title={ERASOR: Egocentric Ratio of Pseudo Occupancy-Based Dynamic Object Removal for Static 3D Point Cloud Map Building}, 
  year={2021},
  volume={6},
  number={2},
  pages={2272-2279},
  doi={10.1109/LRA.2021.3061363}}

@inproceedings{GPR,
title={Gaussian-Process-Based Real-Time Ground Segmentation for Autonomous Land Vehicles[J]},
author={Chen, T. and Dai, B. and Wang, R. and others},
booktitle={Journal of Intelligent and Robotic Systems}, 
year={2014}, 
pages={76(3-4):563-582}
}

@ARTICLE{GraphBasedGS,
  author={Pino, Iván del and Santamaria-Navarro, Angel and Garrell Zulueta, Anaís and Torres, Fernando and Andrade-Cetto, Juan},
  journal={IEEE Transactions on Intelligent Vehicles}, 
  title={Probabilistic Graph-Based Real-Time Ground Segmentation for Urban Robotics}, 
  year={2024},
  volume={9},
  number={5},
  pages={4989-5002},
  doi={10.1109/TIV.2024.3383599}}

@article{Quatro,
author = {Hyungtae Lim and Beomsoo Kim and Daebeom Kim and Eungchang Mason Lee and Hyun Myung},
title ={Quatro++: Robust global registration exploiting ground segmentation for loop closing in LiDAR SLAM},
journal = {The International Journal of Robotics Research},
volume = {43},
number = {5},
pages = {685-715},
year = {2024},
doi = {10.1177/02783649231207654},
URL = {https://doi.org/10.1177/02783649231207654},
eprint = {https://doi.org/10.1177/02783649231207654}
}

@article{TraversabilitySurvey,
  title={Similar but Different: A Survey of Ground Segmentation and Traversability Estimation for Terrestrial Robots},
  author={Lim, Hyungtae and Oh, Minho and Lee, Seungjae and Ahn, Seunguk and Myung, Hyun},
  journal={International Journal of Control, Automation and Systems},
  volume={22},
  number={2},
  pages={347--359},
  year={2024},
  publisher={Springer}
}

@article{tsaiground,
  title={Ground segmentation based point cloud feature extraction for 3D LiDAR SLAM enhancement},
  author={Tsai, Tzu-Cheng and Peng, Chao-Chung},
  journal={Measurement},
  volume={236},
  pages={114890},
  year={2024},
  publisher={Elsevier}
}

@INPROCEEDINGS{PCL,
  author={Rusu, Radu Bogdan and Cousins, Steve},
  booktitle={2011 IEEE International Conference on Robotics and Automation}, 
  title={3D is here: Point Cloud Library (PCL)}, 
  year={2011},
  volume={},
  number={},
  pages={1-4},
  doi={10.1109/ICRA.2011.5980567}}

@article{NANOFLANN,
title={nanoflann: a C++ header-only fork of FLANN, a library for nearest neighbor (NN) with kd-trees},
author={Blanco, Jose Luis and Rai, Pranjal Kumar},
journal={GitHub Repository},
year={2014}
}

\end{document}